\documentclass[letterpaper, 10 pt, conference]{ieeeconf}  

\IEEEoverridecommandlockouts                              

\usepackage{amsmath} 
\usepackage{amssymb}  
\usepackage[utf8]{inputenc}
\usepackage[T1]{fontenc}
\usepackage{hyperref}
\hypersetup{colorlinks=true, allcolors=blue}
\usepackage{graphicx}
\usepackage{comment}
\usepackage[bottom, multiple]{footmisc}
\usepackage{xcolor}
\usepackage{siunitx}
\usepackage{wrapfig}

\usepackage[backend=biber, style=ieee, bibencoding=utf8, sorting=none]{biblatex}

\usepackage{microtype} 
\title{\LARGE \bf
SoFiE: Soft Finger Exoskeleton for Intelligent Grasping
}

\author{Magnus Malthe Sigsgaard Nielsen$^{1}$, Nicklas Nikolaj Grønvall$^{1}$,\\ Xiaofeng Xiong$^{1}$, and Saravana Prashanth Murali Babu$^{1}$*
\thanks{$^{1}$M.M.S. Nielsen, S.P. Murali Babu*, N.N. Grønvall, and X. Xiong  is with SDU Soft Robotics, SDU Biorobotics, The Maersk Mc-Kinney Moller Institute, University of Southern Denmark (SDU), 5230 Odense M, Denmark
        {\tt\small email spmb@mmmi.sdu.dk, magnus.sigsgaard@gmail.com}}%
\thanks{*Authors to whom correspondence should be addressed.}%
}

\begin{document}

\maketitle
\thispagestyle{empty}
\pagestyle{empty}

\begin{abstract}
\noindent Soft wearable robotic systems have emerged as a promising solution for assisting individuals with reduced hand function. This paper presents SoFiE, a modular soft finger exoskeleton designed to assist index-finger flexion during grasping tasks. The proposed system is primarily fabricated using 3D-printed flexible materials, enabling a lightweight, low-profile, and modular design. Actuation is achieved through a tendon-driven mechanism powered by a compact DC motor, while passive extension is provided by a compliant conductive spring. This element, termed StretchSense, also functions as a proprioceptive sensor by exhibiting resistance changes under deformation. Furthermore, a novel tactile sensing approach, MagSense, is introduced, using a magnet–magnetometer pair embedded in a soft fingertip structure to estimate contact force and object compliance. The system is fully untethered and controlled by an embedded microcontroller. In addition, actuator-level sensing through motor encoder feedback enables estimation of the system state, providing a foundation for safe and adaptive control strategies. Experimental validation demonstrates the capability of the system to provide reliable pose estimation, distinguish between materials with different stiffness, and generate distinct sensor signatures across different grasping tasks. This paper details the design, fabrication, and sensing concepts of the proposed exoskeleton as a proof of concept toward modular, soft, and assistive wearable robotics.
\end{abstract}

\section{INTRODUCTION}
\noindent Hand function is central to dexterous movement, enabling grasping, manipulation, self-care, communication, and interaction with the physical environment. When these capabilities are impaired by injury, neurological disorders, or aging, quality of life can be significantly reduced~\cite{Sar_2}. Arthritis-related hand impairment, including hand osteoarthritis and rheumatoid arthritis, is among the most common causes of reduced hand function. These conditions progressively limit hand use through joint pain, stiffness, swelling, reduced range of motion, and loss of grip and pinch strength~\cite{Sar_1}\cite{Sar_2}\cite{Sar_3}. Consequently, daily tasks such as eating, dressing, writing, and opening containers can become difficult and painful. Importantly, arthritis-related hand disability is not only caused by mechanical weakness but also by pain-driven movement avoidance and fatigue during repeated grasping. Clinical studies show that reduced grip strength is strongly associated with ADL limitation and the need for assistive devices in rheumatoid arthritis~\cite{Sar_3}, while hand osteoarthritis is linked to pain, functional decline, and reduced quality of life~\cite{Sar_4}\cite{Sar_5}. Current clinical recommendations include exercises, orthoses, and assistive devices as non-pharmacological strategies to preserve hand function~\cite{Sar_6}. These needs motivate the development of soft finger exoskeletons that can assist grasping, reduce user effort, and provide compliant support without imposing rigid constraints on painful joints.\\ 
Robotic hand exoskeletons have been widely explored for hand rehabilitation and assistance. Conventional systems often use rigid links, mechanical joints, and externally mounted actuation units to transmit forces to the fingers~\cite{Sar_7}. Although these devices can provide precise and repeatable assistance, rigid architectures may introduce joint-misalignment issues, increase mechanical complexity, and reduce comfort during prolonged use~\cite{Sar_8}. Soft robotic hand exoskeletons have therefore gained increasing attention because they use compliant materials and flexible transmission mechanisms that better conform to the human hand~\cite{Sar_9}\cite{Sar_10}. Their intrinsic compliance can improve safety during physical human–robot interaction, reduce the need for precise joint alignment, and support more natural finger motion during grasping. 

\begin{figure}
    \centering
    \includegraphics[width=\linewidth]{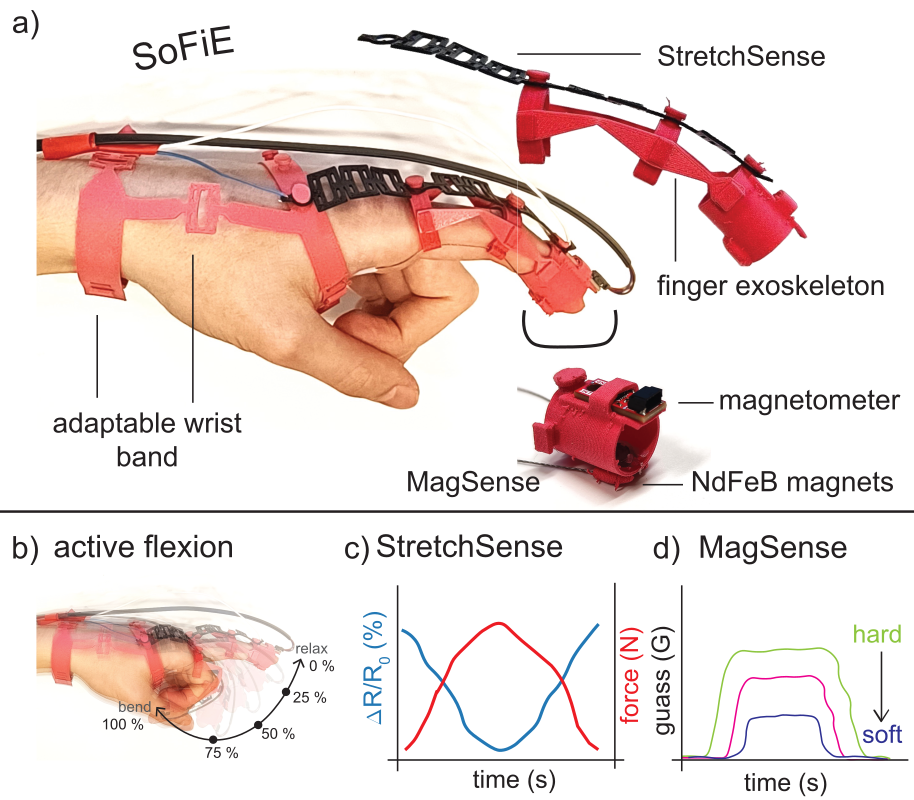}
    \caption{a) SoFiE worn by a user, with key components highlighted, including the StretchSense, MagSense, and adaptable wristband. b) Active finger flexion performed while wearing SoFiE. c) Conceptual illustration of StretchSense behavior in stretched and relaxed states. d) Conceptual illustration of MagSense behavior during contact with objects of different stiffness.}
    \label{fig:Figure1}
\end{figure}

\noindent Several soft glove and hand-exoskeleton systems have demonstrated the potential of this approach. Polygerinos et al. developed a soft robotic glove based on fiber-reinforced elastomeric actuators that were mechanically programmed to support finger bending and grasping, with a portable pressure-control unit for assistance and at-home rehabilitation~\cite{Sar_11}. Yap et al. introduced a fully fabric-based bidirectional soft robotic glove capable of assisting both finger flexion and extension for rehabilitation and task-oriented training~\cite{Sar_12}. Cappello et al. further demonstrated that a fabric-based soft robotic glove could improve object manipulation in individuals with spinal cord injury during activities of daily living~\cite{Sar_13}. Tendon-driven systems have also emerged as compact alternatives to fluidic actuation. Exo-Glove and Exo-Glove Poly showed how soft tendon-routing structures and polymer-based glove architectures can support grasping assistance while reducing rigid mechanical complexity~\cite{Sar_14}\cite{Sar_15}. More recently, soft gloves have integrated sensing, building on the broader potential of magnetic tactile sensing for compact force estimation~\cite{Sar_20}. Other systems have combined textile actuators, soft capacitive sensors, and intent-detection controllers to improve grasping assistance and user interaction~\cite{Sar_16}.\\
Despite these advances, several challenges still limit the translation of soft hand exoskeletons into practical assistive devices. Many systems depend on pneumatic or hydraulic hardware, off-board actuation, or bulky control units, which restrict portability and everyday use~\cite{Sar_8}\cite{Sar_9}\cite{Sar_10}. Other systems improve portability but remain difficult to customize for individual hand anatomy or require complex fabrication and assembly. In addition, sensing is often added as a separate layer rather than being integrated into the mechanical function of the device. By integrating sensing directly into the soft exoskeleton structure, the device can monitor both its own motion and its interaction with the environment without relying on bulky external sensors. Reliable information about finger posture, tendon loading, fingertip contact, and object compliance is essential for safe and adaptive assistance yet achieving this in a compact and wearable platform remains challenging~\cite{Sar_16}\cite{Sar_17}.\\
Modular design and additive manufacturing offer a promising route to address these limitations. By using 3D-printable flexible materials, wearable hand devices can be rapidly customized, repaired, and adapted to different users. Prior work has shown that 3D-printed soft exoskeleton gloves and myoelectric hand orthoses can reduce fabrication complexity and support patient-specific geometries~\cite{Sar_18}\cite{Sar_19}. However, there remains a need for designs that combine soft structure, compact tendon-driven actuation, embedded sensing, and untethered control within a simple and modular architecture.\\
In this work, we present SoFiE, a Soft Finger Exoskeleton for Intelligent Grasping. SoFiE is designed as a proof-of-concept modular, soft wearable system for assisting index finger flexion during grasping. The device is primarily fabricated using flexible 3D-printed components and uses a tendon-driven mechanism actuated by a compact DC motor. Passive return is provided by StretchSense, a compliant conductive spring that also functions as a proprioceptive sensing element through resistance changes during deformation. In addition, we introduce MagSense, a fingertip tactile sensing concept based on a magnet–magnetometer pair to estimate contact force and object compliance. The system is fully untethered and controlled using an embedded microcontroller, while actuator-level sensing from encoder feedback provides additional information about the system state and applied force.

\section{MATERIALS AND METHODS}
\begin{figure}
    \centering
    \includegraphics[width=\linewidth]{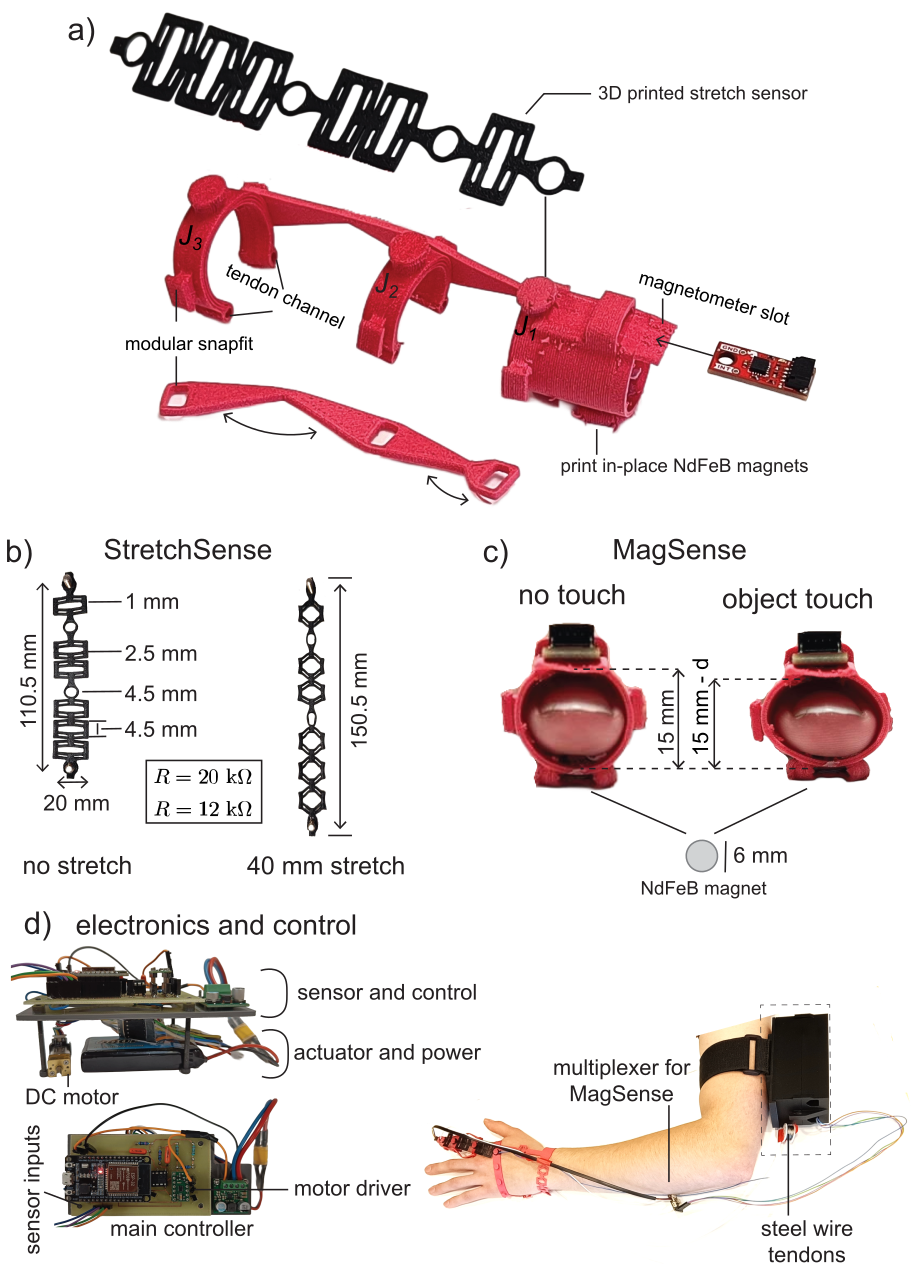}
    \caption{Disassembled view of the index finger module of SoFiE, highlighting the modular design of the exoskeleton. b) StretchSense in the relaxed and fully stretched states, showing a sensing range from \SI{20}{k\ohm} to \SI{12}{k\ohm}. c) Illustration of how finger deformation changes the distance between the magnet and the magnetometer. Starting at \SI{15}{mm} between the magnet and magnetometer, and ending at \SI{15}{mm} minus the distance d. d) Transparent view of the electronic enclosure for the battery and all the electronics for control and actuation, along with an overview of the complete system fitted to the user’s arm and hand.}
    \label{fig:Figure2}
\end{figure}

\subsection{Modular Design}
\noindent The SoFiE exoskeleton is a lightweight, biomimetic system designed to assist finger flexion. The architecture is split into a modular finger glove and an arm-mounted actuation unit, minimizing weight on the hand to reduce fatigue and preserve wrist mobility.
The glove is fabricated via multimaterial FDM 3D printing using ColorFabb VarioShore TPU. By varying the printing temperature, we achieved locally tuned mechanical properties: rigid structural phalanx rings and a soft, porous base for hand conformity. The device attaches via two modular open C-rings and special fingertip unit on the three phalanges (proximal, middle, and distal phalanx) marked as J3, J2, and J1 respectively on Fig.~\ref{fig:Figure2}a. They are  connected by a dorsal compliant spring using snapfit and the achieved modularity allows for rapid adjustment to individual hand geometries, while the flexible TPU negates minor fabrication tolerances (see Fig.~\ref{fig:Figure1}a). The modularity of the exoskeleton is showcased on Fig.~\ref{fig:Figure2}a.

\subsection{Actuation System}
\noindent Finger flexion is driven through a tendon cable mechanism powered by a Pololu 6V Micro Metal Gearmotor (250:1 ratio) with an integrated encoder. A 3 ply stainless steel tendon is routed along the palmar side, looping at the distal phalanx where it returns to a motor mounted spindle. This underactuated system relies on the hand’s natural kinematic constraints to produce coordinated joint motion. Rather than controlling each joint individually, the single tendon routed through the rings creates a natural flexing motion when the motor actuates it. The system is powered by a \SI{7.4}{V} LiPo battery and controlled by an ESP32 microcontroller, housed in a custom enclosure on the upper arm. To ensure stable logic levels, a step-up/down converter maintains a consistent \SI{5}{V} supply from the battery (system overview can be seen on Fig.~\ref{fig:Figure2}d). This fully untethered configuration supports unconstrained movement in diverse grasping scenarios.

\subsection{StretchSense: Integrated Compliant Conductive Sensing}
\noindent A key feature of the design is the integration of StretchSense, a dual-purpose compliant flat spring, connecting the rings at each phalanx as seen on Fig.~\ref{fig:Figure1}a and Fig.~\ref{fig:Figure2}a. StretchSense is fabricated from Recreus Conductive Filaflex TPU (92A Shore hardness). It serves two purposes: providing mechanical passive extension and acting as a strain-dependent resistor for pose estimation.

\subsection{Magsense: Tactile Force Estimation}
\noindent To estimate interaction forces, a small neodymium magnet is embedded in the TPU "pulp" at the distal tip, with a magnetometer (MMC5983MA) mounted dorsally above the fingernail (Fig.~\ref{fig:Figure2}c). This magnet-magnetometer pair uses the TPU and biological tissue as a deformable medium. As contact forces reduce the magnet-to-sensor distance, the resulting shift in magnetic flux density ($B$[G]) is measured. Afterwards it is possible to analyze the relationship $\frac{dB}{dx}$ (flux change over motor displacement) to estimate object stiffness. A second-order Butterworth filter (cutoff: \SI{2}{Hz}) is applied to ensure signal stability.

\subsection{Actuator-Level Sensing and Safety Control}
\noindent By correlating encoder data with StretchSense resistance and MagSense flux, the system achieves robust state estimation. This redundancy allows the controller to identify tendon slack or mechanical obstructions. Software hard-stops based on encoder limits prevent over-extension, while MagSense acts as a secondary safety layer, where a spike in $B$ beyond a threshold triggers an immediate halt. To summarize:
\begin{itemize}
    \item [-] Encoders provide precise data on the displacement on the actuator side.
    \item [-] StretchSense provides a measurement of the exoskeletons deformation and the finger flexion.
    \item [-] MagSense provides confirmation of contact and data about the interaction between the object and the exoskeleton.
\end{itemize}
The multimodal framework provides a foundation for future user-specific control strategies, where the device can adapt its assistance levels based on the wearer's unique grasping mechanics and the compliance of the objects manipulated.

\subsection{Experimental Setups and testing methodology}
\noindent For StretchSense: Five spring thicknesses (\SI{0.5}{mm} to \SI{2.5}{mm}) were tested using a universal testing machine (EZ Test, Shimadzu). To measure the resistive response under strain, an LCR meter (LCR-6100, RS PRO) was interfaced with the ends of the spring via a dedicated test fixture. Data acquisition from the universal testing machine was done through a National Instruments Data Acquisition unit (NI DAQ), which captured the analog force and displacement signals and converted them to digital values. These data streams, along with the digital resistance reading from the LCR meter, were processed and synchronized in real time using a custom MATLAB script. To ensure consistency and eliminate slack, the following experimental procedure was implemented for the 5 different springs: A pre-tension of \textasciitilde\SI{0.15}{N} was applied to the spring. After tensioning, the force and stroke measurements was zeroed to establish a consistent starting datum. Then, each spring underwent 31 extension-relaxation cycles. The springs were stretched to a displacement of \SI{40}{mm} and returned to \SI{0}{mm} at a constant rate of \SI{1}{mm\cdot s^{-1}}. The \SI{40}{mm} stroke was chosen, as that was the maximum stretch measured from the index finger being fully extended to fully contracted. A \SI{5}{s} hold was set at the zero position (\SI{0}{mm}) at the end of each cycle to allow for material contraction.
\begin{figure}[!ht]
    \centering
    \includegraphics[width=\linewidth]{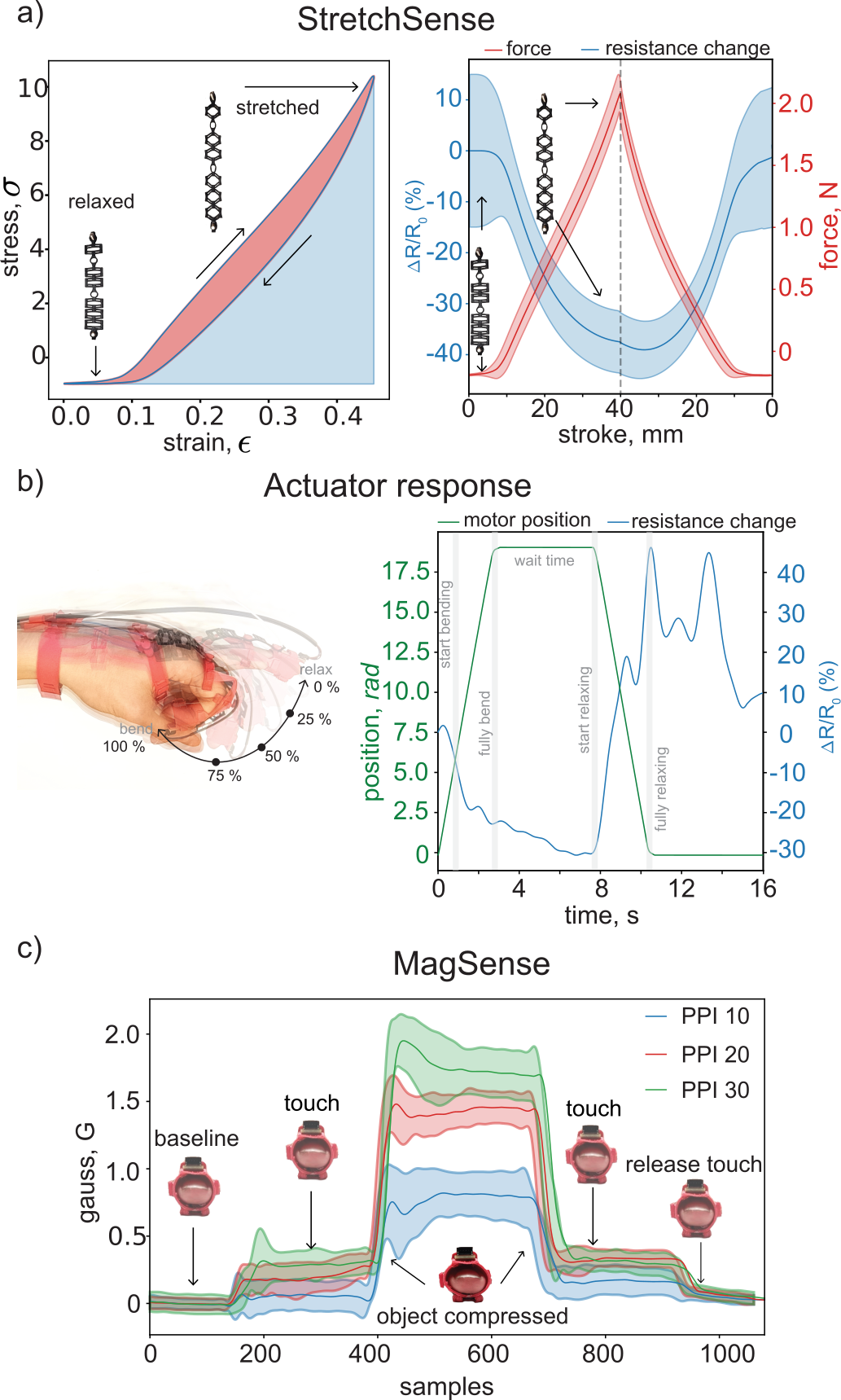}
    \caption{a) left: Stretch strain plot for the \SI{1}{mm} StretchSense showing the physical behavior of the spring. Right: Force and resistive change shown against the stroke distance, showing how the resistance and the force of the spring behave inversely of each other. b) right, As the actuator tightens the tendons and flexes the index finger, the spring is stretched, lowering its resistance. When the tendons are relaxed, the resistance of the spring increases as demonstrated on b) left. c) Response of Magsense when compressing 30, 20, and 10 PPI foam. The firmer the foam, the greater the response of MagSense.}
    \label{fig:Actuator_and_sensor_responses}
\end{figure}
\begin{figure*}[ht!]
    \centering
    \includegraphics[width=\textwidth]{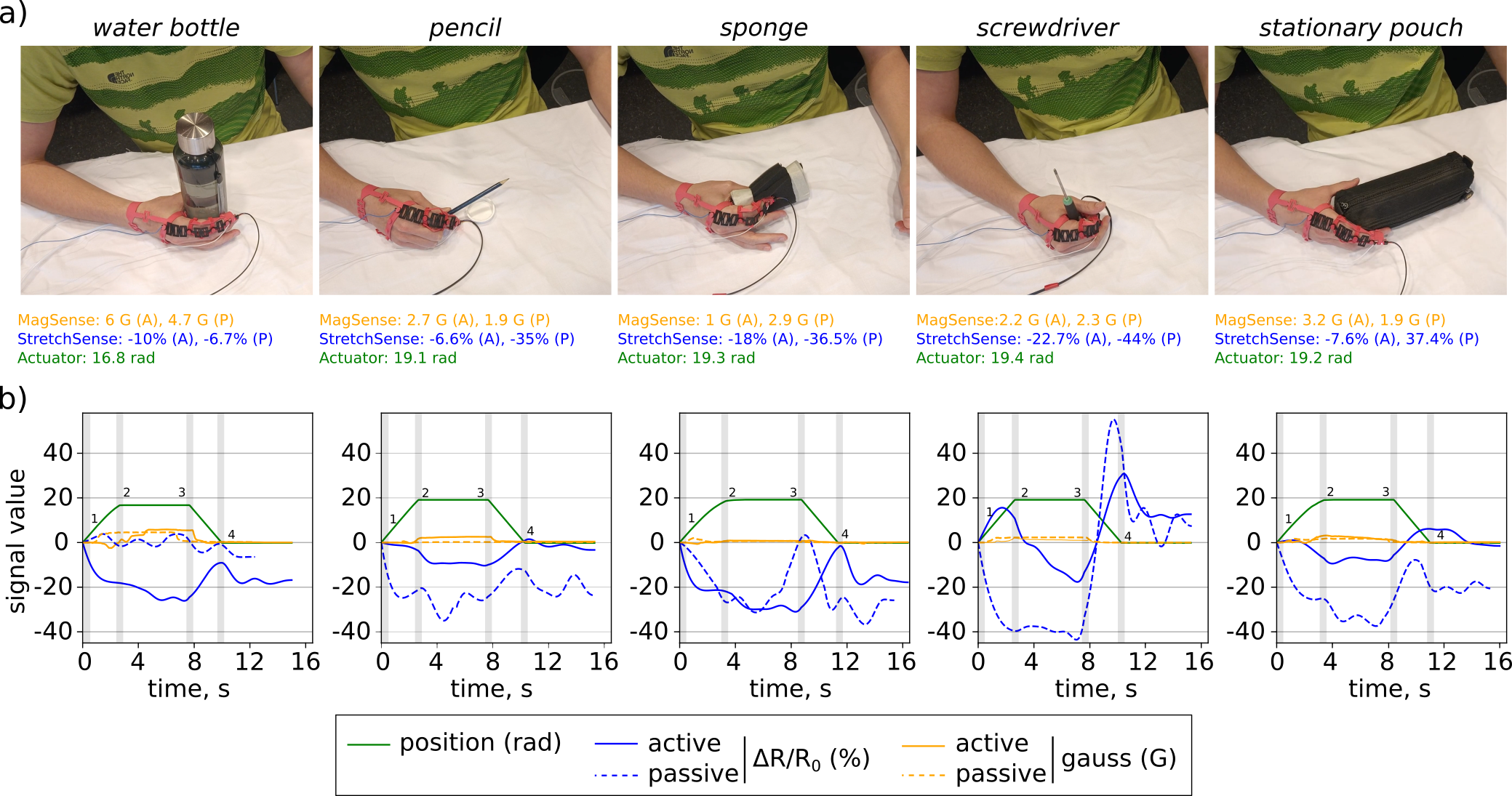}
    \caption{a) Five different everyday objects, measuring the maximum magnetic flux, relative resistance, and position of the motor during grasping for each object. b) Shows the data acquired during grasping for the different objects. On the encoder position, the numbers 1-4 indicates the beginning of actuation, hold, release and rest in that order.}
    \label{fig:figure4}
\end{figure*}
\noindent For post processing, the first cycle of each test was discarded as a "burn-in" outlier to account for initial material settling. The remaining 30 cycles were averaged to determine the mean and standard deviation of the force and resistance as a function of displacement.
To assess long-term performance and transient behavior, the spring selected for final integration was subjected to an extended test. This followed the same aforementioned protocol but was extended to 501 cycles.\\
MagSense was tested against three Polyurethane foams (30, 20, and 10 PPI) of \SI{20}{mm} thickness. A human subject performed 10 trials per foam type, following a 20-second protocol: baseline (\SI{5}{s}), contact with object (hold \SI{5}{s}), \SI{10}{mm} compression (\SI{5}{s}), back to contact (\SI{5}{s}), and then return to baseline.

\section{RESULTS AND DISCUSSION}\label{sec:results}

\subsection{StretchSense}
\noindent The characterization tests showed that all spring thicknesses exhibited a reduction in resistance when stretched. Among the tested designs, the \SI{1}{mm} spring provided the best balance between low restoring force and a wide sensing range of approximately \SI{8}{k\ohm}. Cyclic testing showed that the spring survived 455 cycles before failure. The stress–strain response obtained from fatigue testing indicates that the spring behaves almost linearly during stretching, with a narrow hysteresis band between the stretching and relaxation phases, suggesting similar mechanical behavior in both directions (Fig.~\ref{fig:Actuator_and_sensor_responses}a, left). As shown in Fig.~\ref{fig:Actuator_and_sensor_responses}a, right, the resistance is inversely related to the applied force. This behavior is also observed during active operation of the exoskeleton, as shown in Fig.~\ref{fig:Actuator_and_sensor_responses}b. During actuation, the tendon is tensioned and the finger gradually flexes from the resting position of 0 \% flex to full 100\% flexion, resulting in a decrease in resistance. When the tendon is released, the finger returns toward the resting position of 0\% flex and the resistance increases. Fig.~\ref{fig:Actuator_and_sensor_responses}b, left, shows full finger flexion with the percentages of flexion marked which the plot on Fig.~\ref{fig:Actuator_and_sensor_responses}b, right goes through, demonstrating the range of motion achieved with the exoskeleton. Since the sensor is integrated into the exoskeleton and responds to deformation caused by finger motion, it provides information about the system’s own configuration. It can therefore be regarded as a proprioceptive sensor, as its signal reflects the position and movement of the finger rather than external contact or environmental stimuli.

\noindent A small dip in resistance is observed in Fig.~\ref{fig:Actuator_and_sensor_responses}a, right, as the stroke approaches \SI{0}{mm}. During testing, the spring did not always fully return to its relaxed state before the next stretching cycle, which may explain this behavior. This effect could be related to material fatigue or delayed elastic recovery. However, during normal operation, the spring is typically stretched less than the theoretical \SI{40}{mm} limit; therefore, this effect is unlikely to significantly affect the performance of the exoskeleton during typical use.

\subsection{MagSense}
\noindent The foam tests (Fig.~\ref{fig:Actuator_and_sensor_responses}c) demonstrated that MagSense response increases with material firmness. It can be seen how the baseline is almost 0, until the sensor is lightly placed on the foam where the signal rises by \textasciitilde\SI{0.5}{G}. Then, when the foam is compressed by \SI{10}{mm} the signal increases to \SI{0.7}{G} or up to \SI{2}{G} depending on the firmness of the foam. This validates the magnet-magnetometer pair sensor concept: softer foams (10 PPI) absorb force through internal deformation, whereas the firmer foam (30 PPI) transmit more force to the sensor, bringing the magnet closer to the magnetometer. This distinct flux-to-displacement relationship shows promising behavior for future development into a sensor giving accurate tactile feedback, based on the type of material it is touching, facilitating future adaptive control. Since MagSense is located at the fingertip and responds directly to contact-induced compression, it provides information about external interaction with the environment. It can therefore be regarded as a tactile sensor, as its signal reflects local contact and material-dependent deformation rather than the internal posture of the finger.

\subsection{Demonstration}

\noindent To demonstrate how the StretchSense and MagSense sensors respond during assisted and passive grasping, five demonstrations were performed. Snapshots from the demonstrations are shown in Fig.~\ref{fig:figure4}a. In each demonstration, an everyday object was pinched between the index finger and thumb: a \SI{500}{mL} plastic water bottle, a pencil, a sponge, a screwdriver, and a soft pencil case filled with pencils. These objects represent a range of physical properties. Some are rigid and thin, while others are soft or bulky, requiring different degrees of index finger flexion to achieve a stable grasp.\\Fig.~\ref{fig:figure4}b shows that the sensor responses vary depending on the grasped object. The motion of the finger bending is shown in a sequence (1, 2, 3, 4) from the actuator data. From left to right, the motor position for the water bottle is lower than for the other objects. This is because the bottle is stiff and bulky, allowing a stable grasp to be reached with less tendon displacement. The MagSense response depends strongly on the local contact condition and on how the embedded magnet is pressed against the object. Nevertheless, The MagSense data from both the passive and active exoskeleton in Fig.~\ref{fig:figure4}b show similar trends for the same object, highlighting the potential of the sensor for tactile interaction estimation. A similar behavior is observed for StretchSense. Although the active and passive responses are not identical, they show comparable trends for each object.\\
For example, during grasping of the screwdriver, the finger flexes almost completely around the object, resulting in a larger StretchSense deformation than during the pinching grasps used for the bottle or pencil. The pencil is held mainly between the fingertips, requiring only a small finger motion to achieve a comfortable pinch. The sponge undergoes the largest deformation and is fully compressed when the motor reaches its limit, although the grasp configuration remains similar to that of the pencil. In contrast, the screwdriver requires a more curled grasp, with the index finger wrapping around the object, producing the largest range of finger motion among the tested objects.\\
Finally, the soft pencil case represents an intermediate case. Although the outer structure is soft, the pencils inside provide a semi-rigid interior, placing it between the highly deformable sponge and the rigid objects. For this object, passive grasping produced a larger StretchSense response than active grasping. This difference may be due to variations in how the pencils were arranged inside the case at the moment of contact, introducing variability between trials.

\section{CONCLUSION}
\noindent This work introduced SoFiE, a modular soft finger exoskeleton designed to assist individuals with hand impairments through a lightweight and untethered architecture. By leveraging multimaterial 3D printing and a tendon-driven mechanism, we demonstrated a system that is both customizable and functionally robust. The experimental results validated the effectiveness of a dual-sensing framework for proprioception and tactile feedback.

\begin{itemize}
    \item \textbf{Proprioceptive sensing:} The StretchSense conductive spring successfully provided both passive extension and a measurable resistive response to finger flexion. Although the sensor showed a functional lifetime of approximately 455 cycles, its high-fidelity pose estimation was most effective during larger finger movements.
    
    \item \textbf{Tactile interaction:} The MagSense system was capable of distinguishing between materials with different stiffness. By measuring the displacement-to-flux relationship ($\frac{dB}{dx}$), the sensor identified varying levels of object compliance, from soft foams to rigid tools.
    
    \item \textbf{Sensor fusion:} The demonstration with five everyday objects confirmed that the combination of motor encoders, StretchSense, and MagSense generates distinct data signatures for different grasp types.
\end{itemize}

\noindent Finally, this multilayered sensing approach enables the exoskeleton to infer the mechanical properties of the objects it interacts with. This synergy between soft structure and integrated sensing provides a foundation for future adaptive control strategies, where the device can autonomously modulate assistance levels based on real-time feedback of object stiffness and user intent. Future work will focus on expanding this architecture to a multi-finger system to support more complex functional grasping tasks in daily life.

\addtolength{\textheight}{-7cm}   


\section*{APPENDIX}
\noindent Supporting video can be found at this url: \url{https://youtu.be/rbdGkznkfKY}\\
\noindent Supporting code and 3D files are available on GitHub: \url{https://github.com/SDUSoftRobotics/SoFiE.git}


\printbibliography
\end{document}